\documentclass[10pt,twocolumn,letterpaper]{article}

\usepackage{iccv}
\usepackage{times}
\usepackage{epsfig}
\usepackage{graphicx}
\usepackage{amsmath}
\usepackage{bm}      
\usepackage{amssymb}
\usepackage{pifont}
\newcommand{\cmark}{\ding{51}}%
\newcommand{\xmark}{\ding{55}}%
\usepackage{subcaption}
\usepackage{graphicx}
\usepackage{multicol}
\usepackage{multirow}
\usepackage{enumitem}
\usepackage[skip=0pt]{caption}
\usepackage{soul}

\usepackage[pagebackref=true,breaklinks=true,letterpaper=true,colorlinks,bookmarks=false]{hyperref}

\iccvfinalcopy 

\def\iccvPaperID{10994} 

\ificcvfinal\fi

\begin{document}

\title{PICASO: Permutation-Invariant Cascaded Attentional Set Operator}

\author{Samira Zare,  Hien Van Nguyen\\
Department of Electrical and Computer Engineering\\
University of Houston, Houston, TX, 77004 USA\\
{\tt\small szare836@uh.edu}
}

\maketitle
\ificcvfinal\thispagestyle{empty}\fi

\begin{abstract}
   Set-input deep networks have recently drawn much interest in computer vision and machine learning. This is in part due to the increasing number of important tasks such as meta-learning, clustering, and anomaly detection that are defined on set inputs. These networks must take an arbitrary number of input samples and produce the output invariant to the input set's permutation. Several algorithms have been recently developed to address this urgent need. Our paper analyzes these algorithms using both synthetic and real-world datasets, and shows that they are not effective in dealing with common data variations such as image translation or viewpoint's change. To address this limitation, we propose a permutation-invariant cascaded attentional set operator (PICASO). The gist of PICASO is a cascade of multihead attention blocks with dynamic templates. The proposed operator is a stand-alone module that can be adapted and extended to serve different machine learning tasks. We demonstrate the utilities of PICASO in four diverse scenarios: (i) clustering, (ii) image classification under novel viewpoints, (iii) image anomaly detection, and (iv) state prediction. PICASO increases the SmallNORB image classification accuracy with novel viewpoints by about 10\% points. For set anomaly detection on CelebA dataset, our model improves the areas under ROC and PR curves dataset by about 22\% and 10\%, respectively. For the state prediction on CLEVR dataset, it improves the AP by about 40\%.
\end{abstract}

\section{Introduction}

Many machine learning algorithms are based on fixed dimensional vectors as input data. However, there are a host of tasks that need to deal with inputs or outputs of any arbitrary size. Multiple instance learning \cite{ilse2018attention} is an example of such tasks, where training instances are given as sets, and instead of being individually labeled, one label is provided for the entire set. Alternatively, in some applications, the order of input data should not affect the output. For instance, in 3D shape reconstruction models \cite{yang2020robust}, the input is a set of images and the output needs to be invariant with respect to different permutations of the input images. Recurrent neural networks (RNNs), for example, generate different reconstructions given the same image set but with different permutations \cite{vinyals2015order}. RNNs, therefore, are not appropriate for set representation. Other important applications involving set inputs are point cloud classification \cite{he2020structure}, 3D scene representation by aggregating arbitrary number of objects \cite{eslami2016attend}, cluster assignment or latent variable prediction based on a set of observations \cite{pakman2020attentive, lee2019deep}, and few-shot learning or meta-learning where tasks are specified by a set of arbitrary number of samples \cite{finn2017model, snell2017prototypical,yuan2020few}, and bounding box's aggregation in object detection \cite{ren2015faster,redmon2016you}.  

\begin{figure}[t]
\begin{center}
\includegraphics[width=0.5\textwidth,scale=0.2]{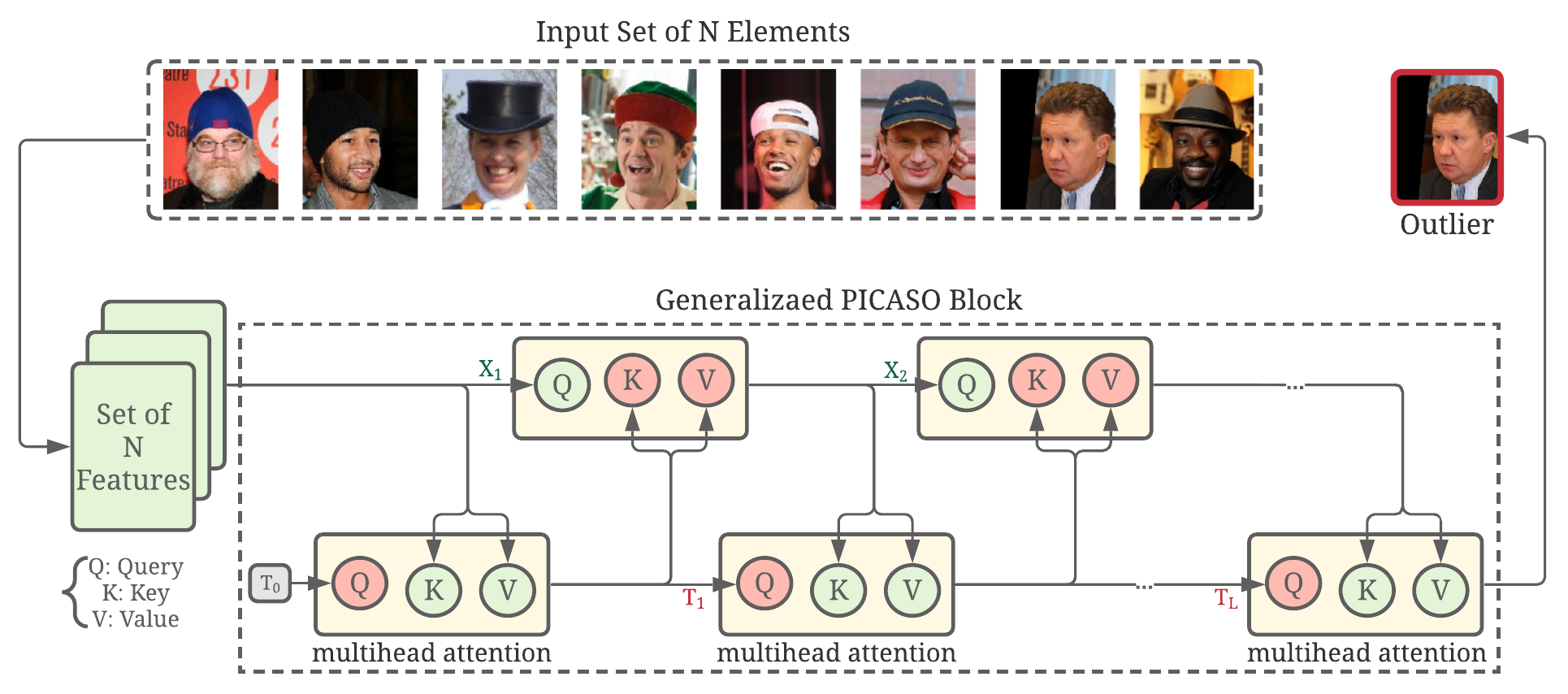} 
\end{center}
   \caption{Set anomaly detection using Generalized version of PICASO. The two attributes smiling and wearing hat are absent in the outlier image. The input set may contain arbitrary number of elements. In generalized PICASO, templates and input set are updated through cascade of multihead attention blocks.} 
\label{fig:onecol}
\end{figure}

In general, these applications require networks that can take inputs with an arbitrary number of samples (a.k.a set inputs) and do not change under any permutations of set members \cite{zaheer2017deep}. The desired permutation-invariant properties of a set-input model are easy to achieve in neural networks. Although there are some recent works \cite{edwards2016towards,zaheer2017deep} to design neural networks for set representation, these models ignore the interactions between set elements and process each element independently. Consequently, the dependencies and relationships between different elements are not learned while they can provide the model with useful information. Another important drawback of these architectures is that they do not have a \emph{learnable} information aggregation process. For example, DeepSets \cite{zaheer2017deep} simply uses max or mean-pooling to gather information across set members. However, we hypothesize that learnable aggregations can make the model more generalizable. Moving from predesigned to learnable information aggregation mechanisms is necessary because our experiments show that the above networks' inference-time performances decrease significantly under common input-set variations.

Set transformer \cite{lee2019set} is a state-of-the-art (SOTA) set-input neural network that can learn to aggregate information across set members using attention mechanisms. This network has shown remarkable performances in working with set inputs and significantly outperformed the approaches with pre-designed information aggregation \cite{edwards2016towards,zaheer2017deep}. The potential capability and simplicity of attention mechanisms (particularly self-attention \cite{vaswani2017attention}) allow this model to attentively distinguish an informative region from a large context while keeping the time complexity of the model low. However, our experiments showed that the set transformer performed poorly when the test data went through common transformations such as image translation or 3D viewpoint change (see Table \ref{tab1}). Besides, we design a clustering experiment where the means of Gaussian Mixtures are shifted during the inference time. Set transformer failed to produce reasonable clustering patterns under this simple generalizability test as shown in Figure \ref{fig2}. More details of our experiments can be found in Section \ref{exp}. These empirical pieces of evidence suggest that there are missing pieces in existing set-input architectures. We hypothesize that the static templates in the set transformer's attention mechanisms prevent it from learning generalizable set operations.


In this paper, we propose a permutation-invariant cascaded attentional set operator, or PICASO in short, with a dynamic information aggregation mechanism. Our operator uses multiple layers of multihead attention blocks to capture higher-order interactions among data points. The central idea is to allow each attention block in the cascade to dynamically form its transformer's templates based on the input data. This operator produces a set of permutation-invariant output vectors each of which can attend to any part in the input. Consequently, PICASO helps the model to learn more general rules and relationships, and better cope with variations in the input data such as image translations or viewpoint changes. PICASO is a multipurpose and simple module that can be used on top of other set-input operators or a CNN encoder to aggregate features attentively. Our paper makes the following contributions:
\begin{itemize}[noitemsep,topsep=0pt]
    \item We propose a novel and effective set operator with dynamic information pooling based on a cascade of transformers. Our transformers' templates are updated based on agreement among data points and the previous templates.
    \item We provide mathematical proofs to show that our cascaded attentional pooling architecture has several desirable properties. Specifically, PICASO is permutation-invariant. Furthermore, its special case is equivalent to the soft K-means algorithm, which is a challenging task for neural networks. 
    \item We validate our operator on a diverse number of tasks, including clustering, image classification under novel 3D viewpoints, set anomaly detection, and state prediction. To facilitate a deeper understanding of the proposed approach, we provide intuitive visualization to show how PICASO attends to different elements of the input set.
\end{itemize}

\begin{table}
\begin{center}
\begin{tabular}{|l|cc|}
\hline
Method & Dynamic Structure & Test Accuracy \\
\hline\hline
DeepSet & \xmark & 68.62\% \\
Set Transformer & \xmark & 75.19\% \\
PICASO (Ours) & \cmark & 85.98\%\\
\hline
\end{tabular}
\end{center}
\caption{Comparison of different set-input approaches on image classification under novel viewpoints using SmallNORB dataset.}
\label{tab1}
\end{table}

\section{Related Works}
\label{others}
\textbf{Pooling operations across set elements:} Different pooling operations have been proposed to aggregate complex features from input sets. In \cite{su2015multi} pooling is utilized across multiple views for 3D shape recognition, while \cite{yang2020robust} and \cite{ilse2018attention} use attention-based weighted sum pooling for multi-view 3D reconstruction and multiple instance learning. \cite{edwards2016towards} use an exchangeable instance pooling to learn the statistics of a set. In \cite{zaheer2017deep} general pooling structures (max, mean, sum, etc.) are discussed and the experiments result in using a permutation invariant sum pooling for problems like estimating the statistics. Janossy pooling \cite{murphy2018janossy} is another permutation invariant approach where the average of a permutation-sensitive function is applied to all reorderings of the input sequence. \cite{murphy2018janossy} show that the method proposed by \cite{zaheer2017deep} is a special case of the Janossy pooling. However, \cite{wagstaff2019limitations} demonstrate the practical limitations of the previous architectures and propose a universal pooling representation for set inputs. Set transformer \cite{lee2019set} uses a permutation invariant pooling method based on multihead attention mechanism following by a self-attention layer. Our work draws inspiration from the set transformer architecture, which learns a fixed template to aggregate the features. In contrast, we design a dynamic attention-based pooling method that updates the template according to the input set. All above-mentioned pooling operations follow a fixed process regardless of the input sets which results in discarding much useful information.

\textbf{Generalization to variations:} In an attempt to achieve translation, rotation, and reflection generalization, \cite{cohen2016group} utilize a CNN-based method with dynamic filters for the object detection task. \cite{worrall2017harmonic} and \cite{esteves2019equivariant} also use CNN-based models with novel convolutional filters that make them equivariant to particular variations. However, these models are not able to deal with variations other than what they are designed for. Capsule networks \cite{sabour2017dynamic, sabour2018matrix} are introduced to learn instantiation parameters and part-whole relationships, which make them viewpoint equivariant. Despite their remarkable performance, capsule networks are hard to train due to the iterative routing algorithm they use \cite{wang2018optimization, li2018neural}. \cite{Ho_2020_CVPR} exploits a self-supervised method using invariant object embedding to make their model generalize to novel viewpoints. In comparison to the existing methods, our model builds templates dynamically in a non-iterative process using an attention mechanism to pool useful features. Further, it can deal with sets with any size as input. New templates are built based on the agreement between input data and a learned fixed template. It helps the model to become more equivariant to variations in the input. The results obtained from the experiments indicate the superiority of our model in generalization to the variations compared to set-function baselines. The details of experiments are discussed in section \ref{exp}.

\textbf{Attention-based models:} Recently, attention mechanisms are broadly used in neural networks for a myriad of applications such as fine-grained classification \cite{xiao2015application, peng2017object} and multi-view 3D reconstruction \cite{yang2020robust}. \cite{mnih2014recurrent} propose a recurrent attention model which can find the location to focus on and process at high resolution. \cite{vinyals2016matching} employ an attention mechanism to pool a weighted average from a learned embedding of the labeled set for one-shot learning and generalization to the test environment. \cite{mishra2017simple} use a combination of self-attention mechanism and temporal convolutions for meta-learning application. \cite{lee2019set} propose an entirely attention-based algorithm for dealing with set input data. \cite{zhang2019self} utilize self-attention layers to help the GAN model understand the dependencies between spatial regions. \cite{locatello2020object} propose some modifications to the dot-product attention and update the slots iteratively. However, it does not take advantage from multihead attention and the interactions between slots.  

\section{PICASO}
\label{method}
In this section, the structure of PICASO and the motivations behind it are described. We introduce an attention-based encoder, used in our experiments, and our proposed attentional pooling operator with its dynamic structure. The main idea is to utilize multihead attention layers \cite{vaswani2017attention} with distinguishing designs to map an input set to particular activations and then pool the useful information through a dynamic structure to generate the outputs. The dynamic attentional pooling operator can extract new templates from the data to aggregate useful information. We also discuss theoretical properties of PICASO such as being a valid set function and equivalence to soft K-means algorithm in the special case. Further, we provide some qualitative analysis of the model through visualization.
\begin{figure}[t]
\begin{center}
\begin{subfigure}{0.2\textwidth}
\captionsetup{justification=centering}
\includegraphics[width=\textwidth,scale=0.1]{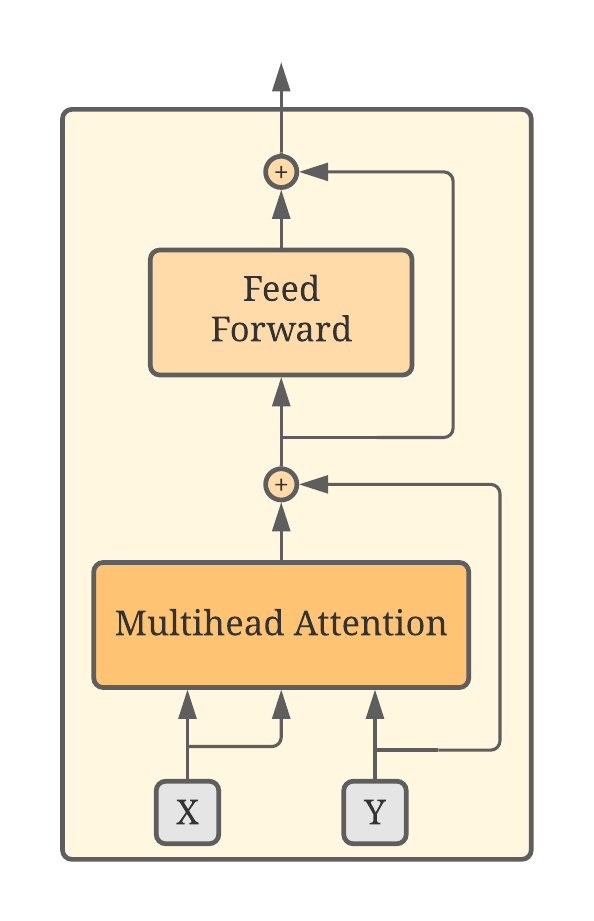}
\caption{Multihead Attention \quad \quad Block }
\label{mab}
\end{subfigure}
\begin{subfigure}{0.2\textwidth}
\captionsetup{justification=centering}
\includegraphics[width=\textwidth,scale=0.1]{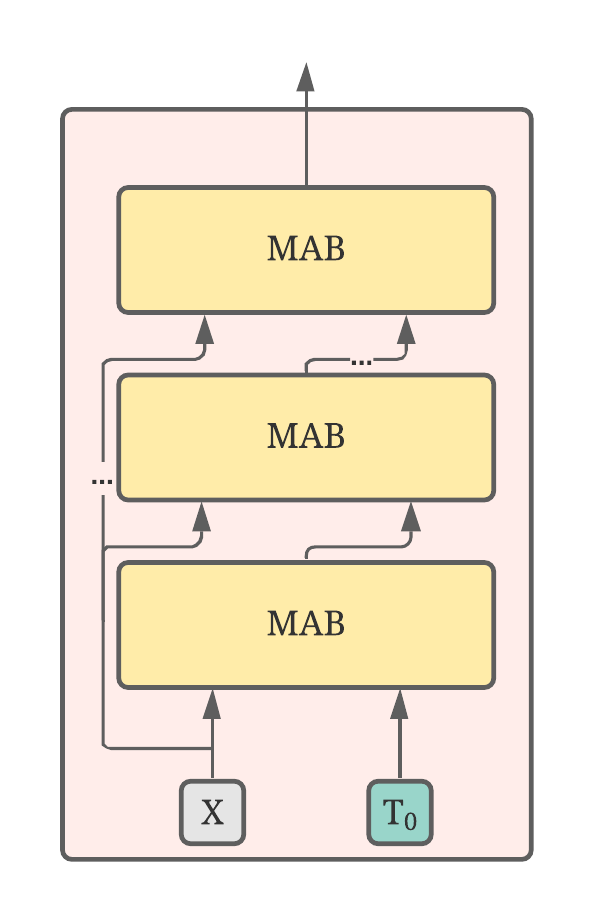}
\caption{PICASO}
\label{dcap}
\end{subfigure}
\end{center}
   \caption{Diagrams of attentional operators}
\label{fig:short}
\end{figure}
\subsection{Transformer and Self-Attention} 
Similar to other variations of the transformer, the core block in this structure is a version of the encoder in the original transformer \cite{vaswani2017attention} that has a multihead attention mechanism with a residual connection followed by a row-wise fully connected feed-forward (rFF) layer with its corresponding residual connection as shown in Figure \ref{mab}. Here we provide a brief introduction to the multihead attention \cite{vaswani2017attention}, self-attention blocks \cite{lee2019set} and our interpretations that are used in our structure. 

Within each head of the multihead attention mechanism, $\bm{Q} \in \mathbb{R}^{n_q\times d_q}$, representing $n_q$ query vectors with dimension $d_q$, builds attention scores based on the similarity between each pair of query and key vectors $\bm{K}\in \mathbb{R}^{n_k\times d_k}$. This similarity is measured by the dot-product of the query with all the keys. Then the value vectors $\bm{V}\in \mathbb{R}^{n_v\times d_v}$ are attended to based on the obtained attention scores. This process can be defined as:
\begin{equation}
\label{eq1}
\text{A(Q, K, V)} = \text{softmax}(\frac{QK^T} {\sqrt{d_k}})V
\end{equation}
Note that Eq.~\ref{eq1} requires $d_k = d_q$ and $n_k = n_v$. The dot-product is divided by ${\sqrt{d_k}}$ to have more stable gradients \cite{vaswani2017attention}. Also, the softmax function normalizes the attention scores along the rows so that they are non-negative and summing to 1. In general, we can have $h$ different heads operating in parallel (a.k.a multihead). Their outputs are concatenated and then linearly transformed to produce the final output. It expands the model's capability to attend to different parts of the input. 
\begin{equation}
\label{eq2}
\begin{split}
\text{Multihead(Q, K, V)} = \text{concat}(h_1, h_2, ... , h_h)W^O \\
 h_j = \text{A}(QW_i^Q, KW_i^K, VW_i^V)
\end{split}
\end{equation}
where $\bm{W}_i^Q$, $\bm{W}_i^K$, $\bm{W}_i^V$, and $\bm{W}^O$ are learnable weights. In the multihead attention structure we are using here, key and value matrices are the same set of $n$ $d$-dimensional vectors: $\bm{X}\in \mathbb{R}^{n\times d}$, and query matrix is a set of $m$ d-dimensional vectors: $\bm{Y}\in \mathbb{R}^{m\times d}$. The multihead attention block (MAB) is defined as follows: 
\begin{equation}
\label{eq3}
\begin{split}
\text{MAB(Y, X)} = H + rFF(H) \\
 H = YW^Y + rFF(\text{Multihead}(Y, X, X))
\end{split}
\end{equation}
The block-diagram of MAB is shown in Figure \ref{mab}. MAB determines the interactions and dependencies between the queries and keys regardless of their orders in the set and computes attention scores. In other words, the query vectors determine which values to attend. Therefore, query vectors can be seen as templates, which guide the model to selectively pay attention to different values. It is worth noting that in a special case where $\bm{X}$ and $\bm{Y}$ are the same, MAB is equivalent to a well-known multihead self-attention mechanism \cite{vaswani2017attention} whose output includes information about interactions among different elements in the input set.

Although transformers reduce the computational complexity per layer in comparison to recurrent or convolutional layers, due to the pairwise interactions in the attention mechanism, the complexity of self-attention mechanism is $O(n^2)$, which can be costly for large sets. In order to reduce the time-complexity, \cite{lee2019set} introduce an induced version of the self-attention (SA) mechanism, which contains a learnable matrix, named inducing points. Within an intermediate step, the input set $\bm{X}\in \mathbb{R}^{n\times d}$ is attended to and transformed into a low dimensional matrix $\bm{Z}\in \mathbb{R}^{k\times d}$ using the inducing points as the query matrix in the multihead attention mechanism. Then $Z$ is attended to by the input set to produce outputs and extract useful features: 
\begin{equation}
\label{eqa}
\begin{split}
 Z = \text{MAB}(T, X)\in \mathbb{R}^{k\times d}\\
 X' = \text{MAB}(X, Z)\in \mathbb{R}^{n\times d}
\end{split}
\end{equation}
where $k$ is the number of learnable templates $\bm{T}\in \mathbb{R}^{k\times d}$ depending on the task of interest. The architecture is more computationally efficient than the original transformer ($O(kn)$ versus $O(n^2)$, where $k \ll n$). The two steps resemble to an autoencoder that maps the input to a lower dimension within an intermediate step and reconstructs it again in a higher-dimensional space. This process can be repeated multiple times to build the attentional encoder (AE) overall architecture. Each individual AE block first extracts a set of features from the input according to a learned template (fixed at inference time), then uses these features to create attention scores and build the attentional version of the input. 

\subsection{PICASO Block} 
Transformer can also be used to attentively summarize the input set into a fixed-length object using a learnable set of $k$ $d$-dimensional query vectors $\bm{Q}\in \mathbb{R}^{k\times d}$. As a result, the output will be a set of $k$ $d$-dimensional points. The pooling operator in set transformer architecture, known as PMA, aggregates the features as a weighted sum where the weights are computed by attending to the input features. However, the problem with this block is that it uses one learnable but static templates (i.e, independent of the input data at inference time), which keep the model from dynamically adapting to the input set. Hence, this model will act like a traditional CNN, where filters are fixed. But we aim to create dynamic templates to let the model extract richer information related to the data, and consequently, increase the capability of the model. 

We propose a dynamic attentional pooling method, named PICASO block (PB), that updates its template depending on the previous template and its agreement with the encoded input data through consecutive multihead attention blocks. Again, each template is made up of a set of $k$ components, i.e. $\bm{T}\in \mathbb{R}^{k\times d}$, where $k$ can be selected based on the task. For example, we select $k=c$ for a clustering problem with $c$ clusters, and $k=1$ for classification task. In contrast to existing approaches, our templates are updated based on the previous ones using a multihead attention mechanism starting from fixed learned templates $T_0$. The block diagram is shown in Figure \ref{dcap} and it can be defined as:
\begin{equation}
\label{eq9}
\begin{split}
T_{i+1} = \text{MAB}(T_i, X)\in \mathbb{R}^{k\times d}, \quad i=0, 1 \ldots, L-1
\end{split}
\end{equation}
Note that PMA in the set transformer is a special case of PICASO where the template is not updating and $i=0$.

\noindent \textbf{Generalized PICASO Block (GPB).} In specific tasks where we may need a more complex model, we can update $X$ as well. This way, we reconstruct the set based on the dynamically aggregated features. 
\begin{equation}
\label{eqg}
\begin{split}
T_{i+1} = \text{MAB}(T_i, X_i)\in \mathbb{R}^{k\times d} \\
 X_i = \text{MAB}(X_{i-1}, T_{i})\in \mathbb{R}^{n\times d}
\end{split}
\end{equation}
As can be seen in Eq. \ref{eqa} and Eq. \ref{eqg}, generalized PICASO can also be considered as a version of AE with an extra connection that makes the new template. However, in AE we use learnable static templates and get $X_i\in \mathbb{R}^{n\times d}$ as the output, which contains element-wise features. On the other hand, the output of generalized PICASO is $T_{L}\in \mathbb{R}^{k\times d}$, which represents the population-wise aggregation (i.e. aggregates all input features).

Another point to be mentioned is that parameters are shared among all steps of the updates to avoid over parametrization. The overall time-complexity of the module is $O(i\times n\times k)$. PICASO maps a set of $n$ input feature vectors into a set of $k$ template vectors by iterating through $i$ steps. In each step, templates update their representation via an attention mechanism to better explain the input from the previous step. PICASO module can be used on top of a cascade of AEs, CNN or any other encoder, and its final representation can be passed to the output layer or a downstream task. Here, we discuss some properties of PICASO theoretically.



\begin{figure*}[t]
\begin{center}
   \includegraphics[width=0.85\linewidth, scale = 0.95]{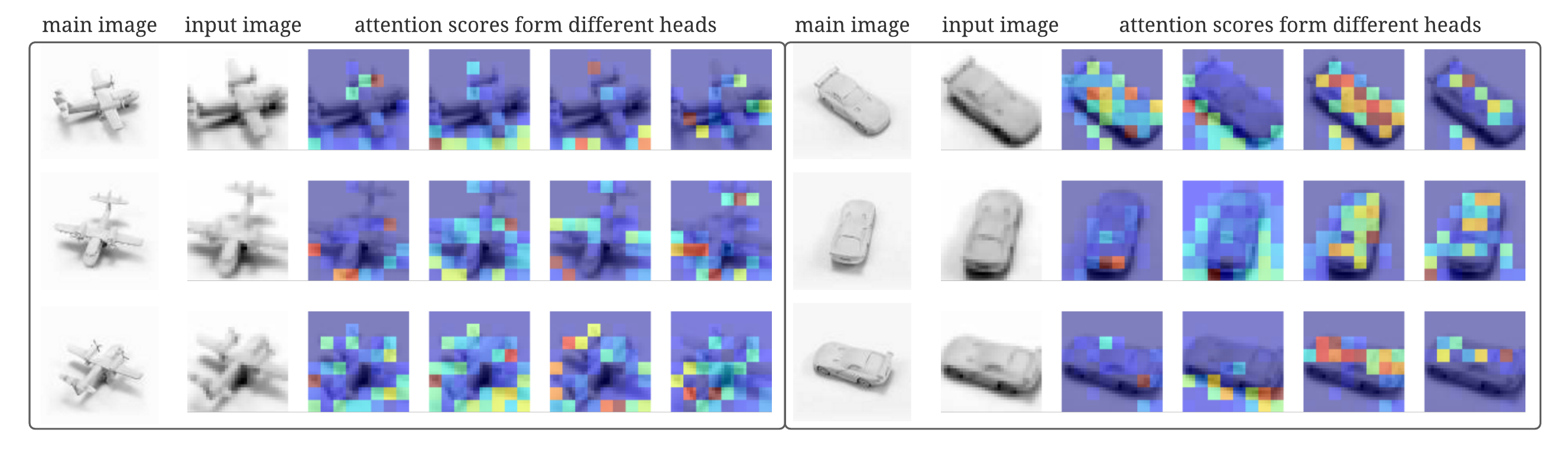}
\end{center}
   \caption{Importance visualization of set members for smallNORB classification with novel viewpoints}
\label{vis}
\end{figure*}

\textbf{Definition 1.} A set operator is called permutation-invariant iff any permutation of set elements, $\pi$, does not change the output, \ie $f(\pi x) = f(x)$

\textbf{Proposition 1.} PICASO is a valid set function, meaning that (i) it can have inputs of arbitrary size, (ii) it is permutation-invariant.

\textit{Proof.} (i) To show that the model has the first property, consider an attention block where $\bm{T}\in \mathbb{R}^{k\times d_t}$ and $\bm{X}\in \mathbb{R}^{n\times d_x}$. Correspondingly, we would have $\bm{W^T}\in \mathbb{R}^{d_t\times d}$ and $\bm{W^X}\in \mathbb{R}^{d_x\times d}$. Therefore, weight matrix dimensions do not depend on $n$ and $k$. In this case, it can be seen easily that the output of an MAB is $k \times d$. 

(ii) The second property is also achieved considering the fact that each elements of an attention block output is a weighted sum of values. The $ij$-th element of the output can be represented as follows, where $x_i , t_m\in \mathbb{R}^{1\times d}$:
\begin{equation}
\label{eq11}
\begin{split}
\text{A(T, X, X)}_{ij} =  \begin{bmatrix} \sum_{m=1}^{n} \text{softmax}(\frac{t_i.x_m} {\sqrt{d_k}})x_{mj} \end{bmatrix}
\end{split}
\end{equation}
If the order of set elements changes, the order of attention scores will change correspondingly and the output of the attention block will not change. Therefore, the model is permutation invariant.

\textbf{Proposition 2.} A special case of the PICASO operating on a set $X$ can represent the soft K-means algorithm.

\textit{Proof.} In the K-means algorithm, we start with an initial guess for the centroids, and then find the distance between each point and every individual center using any kind of distance. Next, points are assigned to the cluster center with minimum distance. Then, centroids are updated by calculating the mean of the new assigned clusters. Again, the distances are found and the whole process is repeated until no data point is reassigned.  

Now, consider the problem of clustering a dataset containing $n$ points from $k$ components. Without loss of generality, suppose we do not use weight matrices in Eq. \ref{eq2}. Knowing that each row of the template $\bm{T}$ represents one cluster and $t_i , x_j\in \mathbb{R}^{1\times d}$, we would have:
\begin{equation}
\label{eq10}
\begin{split}
\text{A(T, X, X)} = \text{softmax}(\frac{TX^T} {\sqrt{d}})X \\= 
\text{softmax}(\frac{1} {\sqrt{d}} \begin{bmatrix} t_{1}.x_1 & t_1.x_2 & ... & t_1.x_n\\ \vdots &\vdots &\vdots & \vdots\\ t_{k}.x_1 & t_k.x_2 & ... & t_k.x_n\end{bmatrix})X \\
\end{split}
\end{equation}

The dot-product operation measures the cosine distance between each point and each initially placed centroid represented by the template, as required in the K-means algorithm. Assuming that the output of the softmax function would produce the probability of assigning each point to a particular cluster or a set of assignment vectors based on their agreement, we will have:
\begin{equation}
\label{eq11}
\begin{split}
\text{A(T, X, X)}_{ij} = \begin{bmatrix} \sum_{m=1}^{n} p(x_m|c_i)x_{mj} \end{bmatrix}
\end{split}
\end{equation}

Consequently, A(.) is a representation of the centroids. Therefore, in the special case where there is no residual connection, the proposed PICASO can be interpreted as a version of soft K-means clustering method considering the assignment probability of data points with $T_0$ as the initial guess. Similar to the K-means algorithm, the calculated mean in each step is used in its following step to find the updated mean. Our clustering results on synthetic Gaussian mixtures dataset in Section~\ref{exp} confirm this observation. Also, we found that using residual connections and feed-forward layer can help improve performance. 

\subsection{Importance Visualization of Set Members}
In order to have a qualitative analysis of the algorithm, we visualize the attention weights that are computed in PICASO using dynamic templates. Each template is attending to a set of $n$ elements and we have an attention weights for each member obtained from Eq. \ref{eq1}. Further, each multihead attention block has $h$ separate heads, allowing the model to attend to input elements with various weights. Thus for importance visualization of set elements, we visualize weights assigned to each member for different heads. Note that outputs from different heads are then concatenated to compute the output. In Figure \ref{vis}, the attention weights in each head from the last updating step for smallNORB classification experiment are shown. In this experiment, each input image is represented as a set of 64 4$\times$4 squares. To achieve a better interpretation, we visualize the weights for different viewpoints of a single object, which is classified correctly. Each square (4$\times$4 pixels) corresponds to a set member and the heatmap represents the score magnitude. Looking at the airplane from different viewpoints, for example, the model learned to locate and attend to the body shape and wings to classify the image.

\section{Experiments}
\label{exp}

We evaluate the PICASO module on different tasks--clustering, classification, and anomaly detection scenarios. We compare against specialized SOTA methods for each respective task. The code is available at \url{https://github.com/samzare/PICASO}.

\textbf{Datasets.} For the clustering experiments, we use a synthetic 2D mixture of Gaussians. We generate input sets of arbitrary $n$ 2D points $\bm{X}\in[x_{min}, x_{max}]$ sampled from a Gaussian mixture with 4 different components. At the inference time, the model performance is evaluated on shifted data points. For classification task, we use the smallNORB dataset \cite{lecun2004learning}, which is designed specifically for 3D object recognition. It contains gray-scale images of 5 generic categories of toys including car, airplane, animal, humans, and trucks and each class has 10 different instances. Every single object is captured using 2 cameras at 9 elevations (30-70 degree), 18 different azimuth (0-340 degree), and 6 lighting conditions. Similar to \cite{sabour2018matrix}, we use one-third of the azimuths and elevations for training and evaluate the model on novel viewpoints. Further, for the set anomaly detection experiments, we utilize CelebA dataset \cite{liu2015faceattributes} composed of more than 200K face images with 40 different attributes. Similar to \cite{lee2019set}, we build 1000 sets of randomly sampled images. Each set contains seven images with two randomly selected attributes and one target image with neither. The goal is to find the target image with both attributes absent in each set. For the state prediction experiment, we use CLEVR dataset \cite{johnson2017clevr} with 100K images and its original train/test split. Images contain three to ten objects with different properties (position, shape, material, color, and size) and the network is trained supervisedly to predict object properties. 

\textbf{Baselines.} We compare the performance of our algorithm with two other widely known set-based networks as baselines: DeepSet \cite{zaheer2017deep}, and Set Transformer \cite{lee2019set}. DeepSet uses mean or max pooling on set elements and Set Transformer has a static attentional pooling. For the DeepSet baseline, we use their permutation invariant model in our experiments, and for the Set Transformer baseline, we compare with the published settings in \cite{lee2019set} and other variations of their architecture. For the classification task, the results are also compared with CapsNet \cite{sabour2018matrix} as a SOTA model. CapsNet has an iterative dynamic structure that helps it to generalize to novel data variations. For the state prediction, results are compared against Deep Set Prediction Networks (DSPN) \cite{zhang2019deep} and Slot Attention \cite{locatello2020object}. For the CapsNet, DSPN, and Slot Attention baselines, we compare with their published results since we use the same experimental setup.
\begin{table*}
\begin{center}
\begin{tabular}{|l|ccccccc|}
\hline
\multirow{2}{4em}{\bf Architecture} & \multicolumn{7}{c|}{\bf Log-Likelihood when the mean is shifted for}\\
& \bf 0 & \bf +8 &\bf -8 &\bf +10 &\bf -10  &\bf +12 &\bf -12\\
\hline\hline
DeepSet  &-2.1554 &-9.2896&-15.6981&-11.7255&-18.4507&-14.1290&-23.5431\\
Set Transformer    & -1.5634 &-2.2196 &-2.2928 &-2.5896 &-2.8368 &-2.9949 &-3.6319\\
AE$_{16}$ + PMA     &  -1.5700 &-2.1487 &-2.3363 &-2.6069 &-3.0373 &-3.1790
&-3.8504\\ 
AE$_{32}$ + PMA & -1.5529 &-2.2316 &-2.4960 &-2.5145 &-2.9975 &-2.8237 &-3.5220\\
\hline
AE$_{16}$ + PB & -1.5892 &-2.0492 &-2.3424 &-2.3261 &-2.8485 &-2.5434 & -3.4351\\ 
SA + PB & -1.6327 &-2.6569 &-2.4408 &-3.3984 &-2.8037 &-4.5694 &-3.1316\\

AE$_{32}$ + PB & -1.5191 &-2.4391 &-2.0305 &-2.9861 &-2.2699 &-3.4665 &\textbf{-2.6099}\\
AE$_{32}$ + PB + SA & \textbf{-1.5020} & \textbf{-1.8994} &\textbf{-1.9528} & \textbf{-2.1654} &\textbf{-2.1838} &\textbf{-2.4921} &-2.6190\\
\hline
\end{tabular}
\end{center}
\caption{The average log-likelihood per data for 1000 sets of synthetic Gaussian mixtures along with their correspondence shifted sets are shown for different architectures. AE$_{m}$ shows the number of inducing points. The oracle loss (the log-likelihood using the parameters used to generate the dataset) is $-1.4697$ for all experiments.}
\label{MOG}
\end{table*}

\subsection{Clustering of 2D Mixture of Gaussian}
\begin{figure}[t]
\begin{center}
\begin{subfigure}{0.21\textwidth}
\includegraphics[width=\textwidth,scale=0.1]{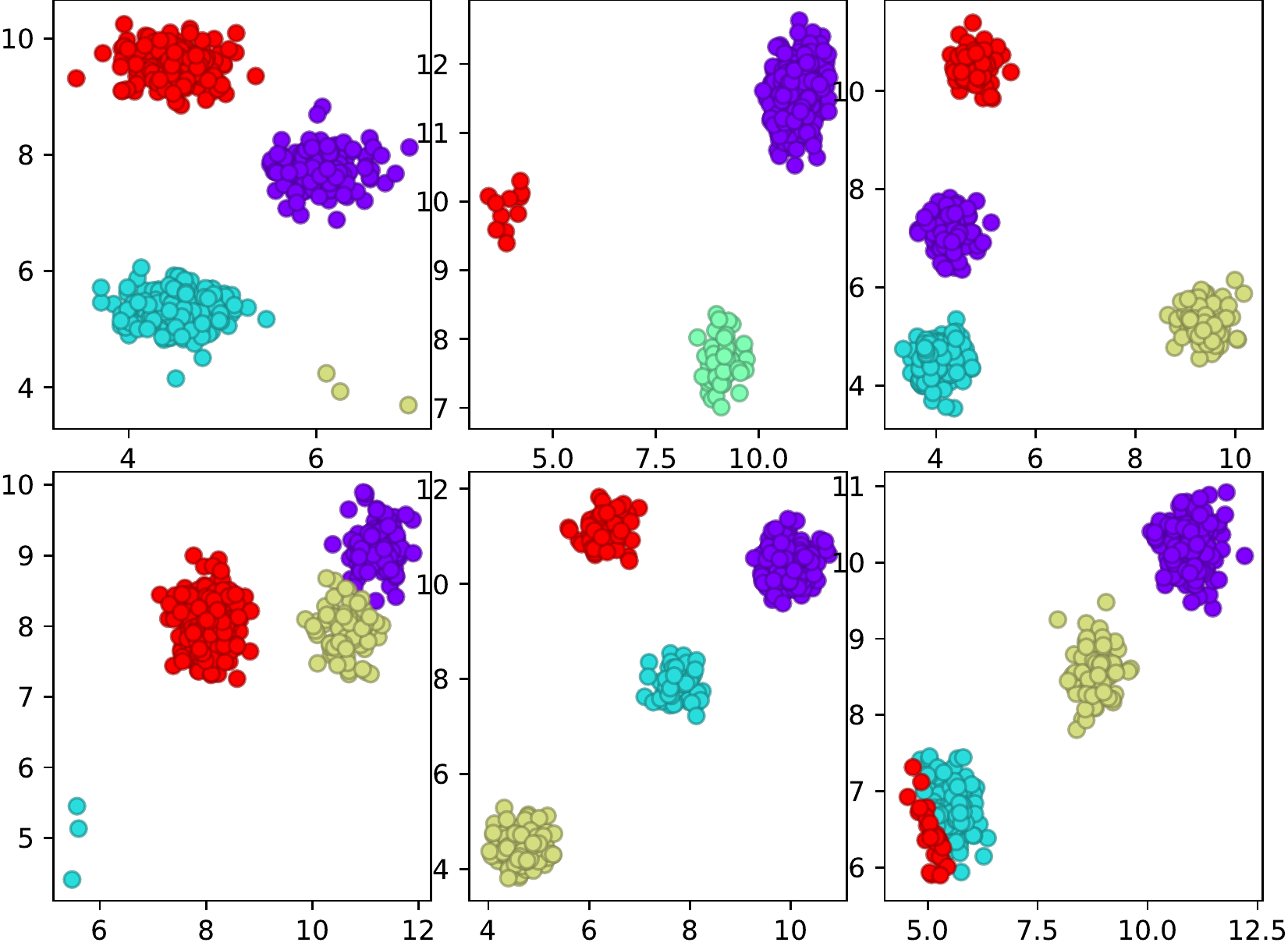}
\label{fig:subim2}
\end{subfigure}
\begin{subfigure}{0.21\textwidth}
\includegraphics[width=\textwidth,scale=0.1]{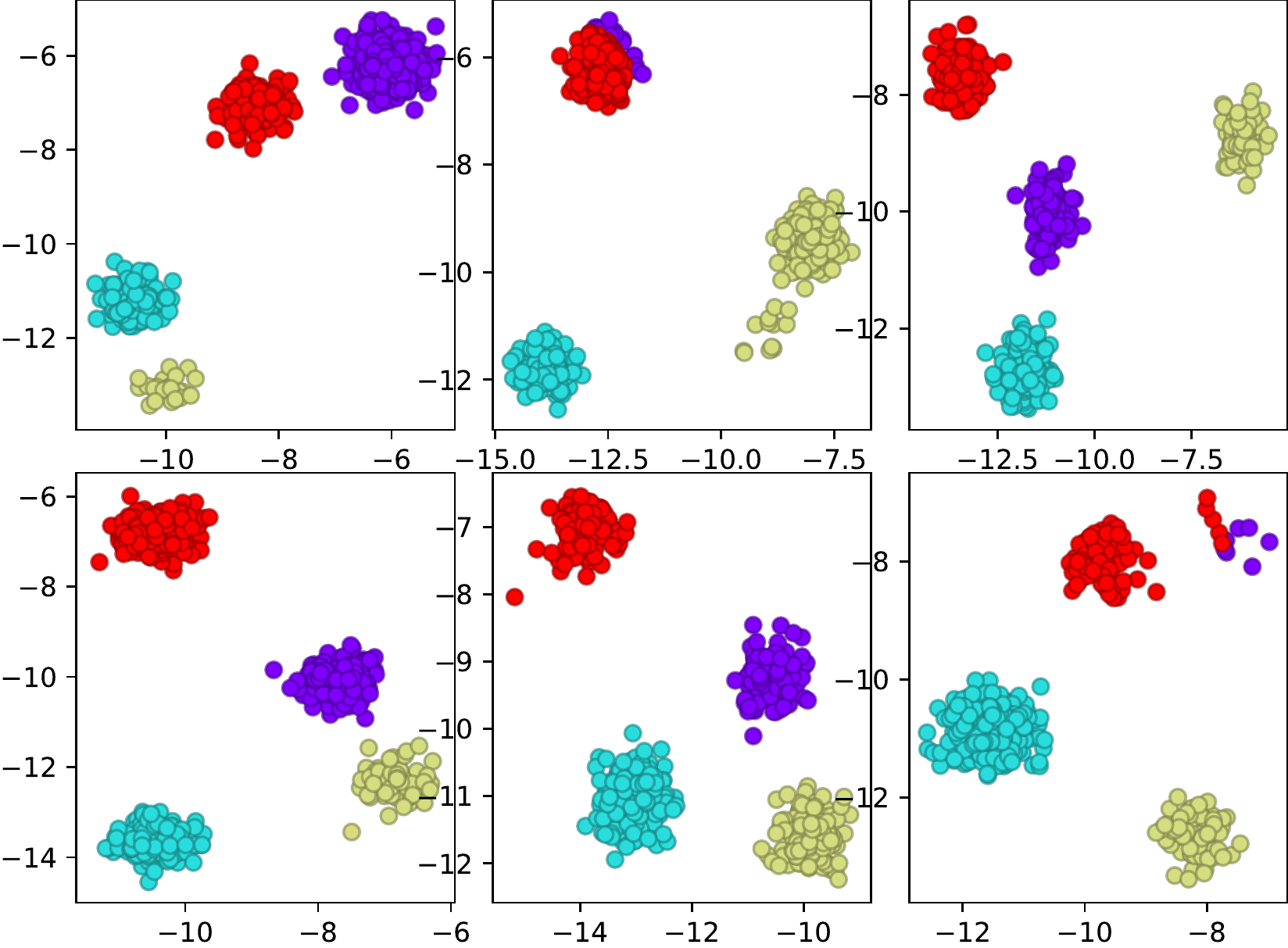}
\label{fig:subim3}
\end{subfigure}

\begin{subfigure}{0.21\textwidth}
\includegraphics[width=\textwidth,scale=0.1]{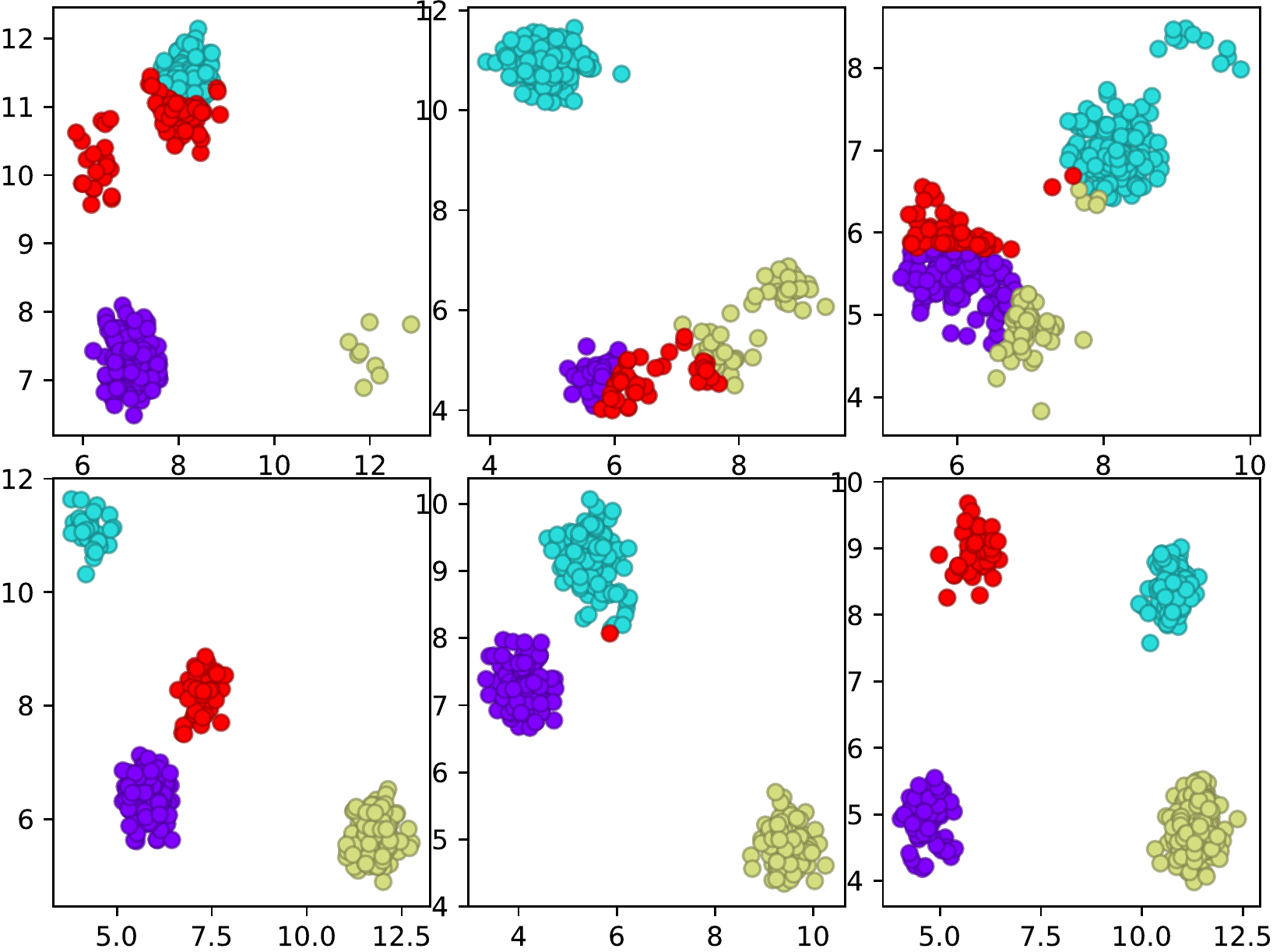}
\caption{shift = +8}
\label{fig:subim2}
\end{subfigure}
\begin{subfigure}{0.21\textwidth}
\includegraphics[width=\textwidth,scale=0.1]{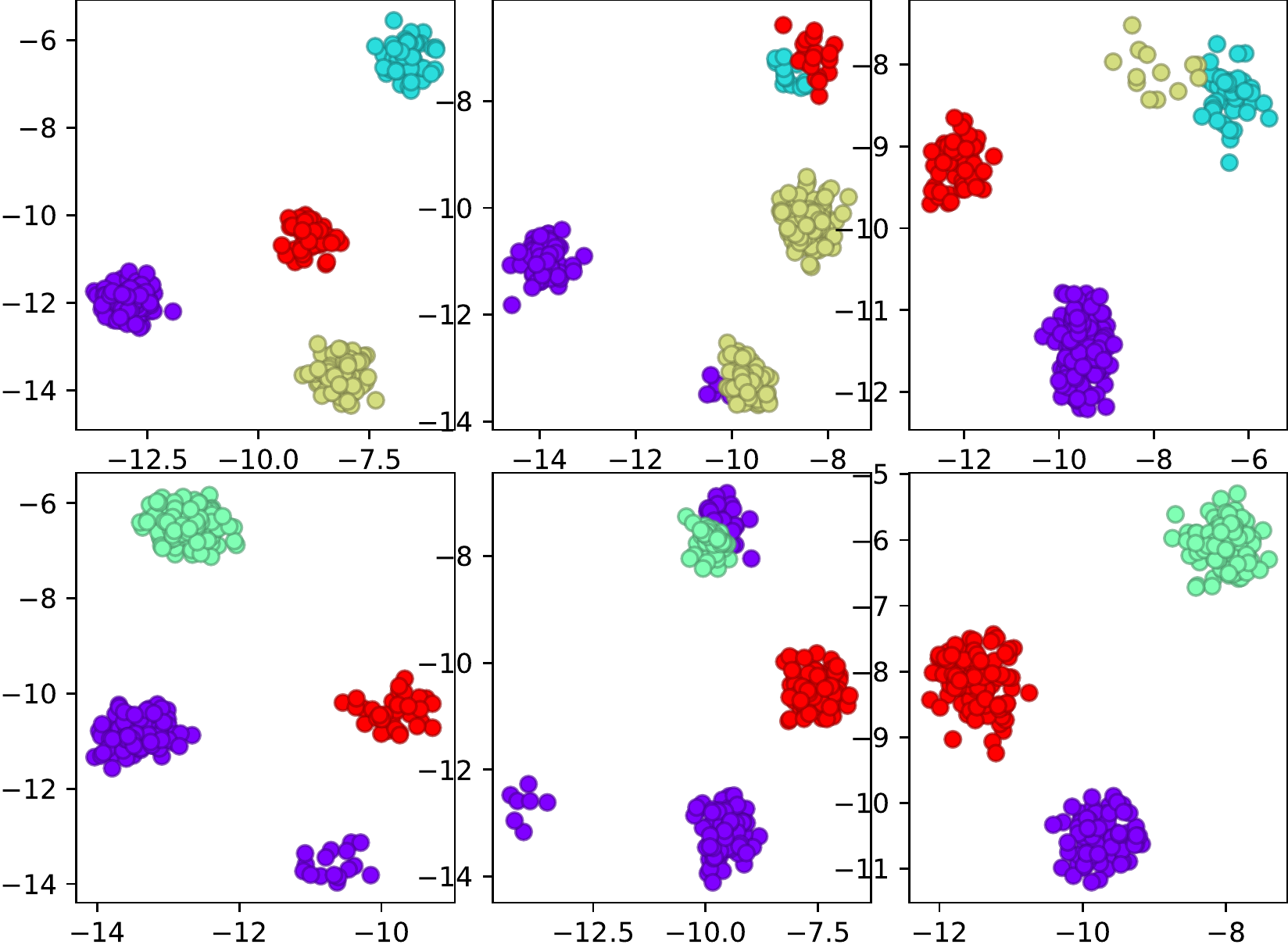}
\caption{shift = -10}
\label{fig:subim3}
\end{subfigure}
\caption{Clustering results for 6 test datasets with different shifting values. PICASO (top), Set Transformer (bottom).}
\label{fig2}
\end{center}
\end{figure}

First, we apply PICASO on clustering the Gaussian mixture with $k=4$ components and an arbitrary number of points $n\in[300, 600]$. The mean of each component is selected randomly $\mu_j\in[-4, 4]$ and the variance is assumed as $\sigma_{j}=0.3$. This model directly learns the parameters $\hat{\theta}(X)=[\{\mu_j, \sigma_j\}^k_j{}_={}_1, \pi(X)]$ so that we have maximum log-likelihood. The model is trained using stochastic gradient ascent to maximize the log-likelihood. Log-likelihood function for a dataset $X$ generated from a Gaussian mixture with $k$ components is as follows:
\begin{equation}
\label{eq8}
\begin{split}
\mathcal{L}=\mathbb{E}\big[\sum_{i=1}^{n} log \sum_{j=1}^{k} \pi_j(X)\mathcal{N} ( x_i ; \mu_j(X) , \Sigma_j )\big]
\end{split}
\end{equation}
To evaluate the ability of PICASO in dealing with changes in the input set, the performance of the trained model is evaluated using data points shifted by $0, \pm 8, \pm 10, \pm 12$ so that the set elements are moved out of the range of training data. We use learnable templates $\bm{T}\in \mathbb{R}^{4\times d}$ along the network and each row of them $t_j\in \mathbb{R}^{1\times d}$ will be responsible for estimating the characteristics of a cluster. In this experiment, PB is used on top of SA or AE blocks with 16 or 32 inducing points and is made up of two steps of updating with weight sharing. Therefore, the number of parameters is not increased compared to the Set Transformer baseline. We also compare our performances to the architecture proposed in \cite{lee2019set} and its variants. The results in Table \ref{MOG} and Figure \ref{fig2} indicate that our module can help the model to better cope with the variations of input sets compared to the baselines. Through its updating steps, PICASO can learn the dependencies between data points by finding agreements that relate them.

\subsection{SmallNorb Classification}
For the classification experiments, we train the model on one-third of smallNORB training images containing azimuths of (300, 320, 340, 0, 20, 40) and test on test data that contains different azimuths. In another experiment, the model is trained on elevations of (0, 1, 2) and tested on other elevations. In both, each image is downsampled to 48$\times$48 pixels, and then, a 32$\times$32 patch is cropped. To make sets out of every individual image, each image is divided to 64 squares of size 4$\times$4. Then, each square is flattened to make it possible to be used as inputs of our model. In other words, each image is represented as a set of 64 16D elements. The models are trained using the Adam optimizer \cite{kingma2014adam} with a learning rate of $1 \times 10^{-4}$ and a batch size of 128. We use the same training setting across all models.
We evaluate the classification accuracy for different number of updating steps (PB$_{i}$) with shared weights. Hence, the number of parameters is approximately the same as Set Transformer baseline. Results, listed in Table \ref{table3}, shows our model outperforming SOTA set deep networks. This observation indicates that dynamic template update coupled with an attention mechanism is more effective in capturing object-part relationships. The CapsNet with EM routing \cite{sabour2018matrix} reported 86.5\% and 87.7\% accuracy for novel azimuth and elevation, respectively, which are close to our results. 
\begin{table}
\begin{center}
\begin{tabular}{|l|cc|}
\hline
   & \bf Novel Azimuth &\bf Novel Elevation \\
\hline\hline
DeepSet       &68.62\%   &64.10\%\\
Set Transformer    &75.19\%  &74.26\%\\
\hline
AE$_{}$ + PB$_{1}$       &79.36\%  &78.69\%\\
AE$_{}$ + PB$_{2}$       &83.38\%  &82.18\%\\
AE$_{}$ + PB$_{3}$       &85.98\%  &84.47\%\\
\hline
\end{tabular}
\end{center}
\caption{SmallNORB test results with novel viewpoints}
\label{table3}
\end{table}
\begin{figure}
    \centering
    \includegraphics[width=0.5\textwidth,scale=0.1]{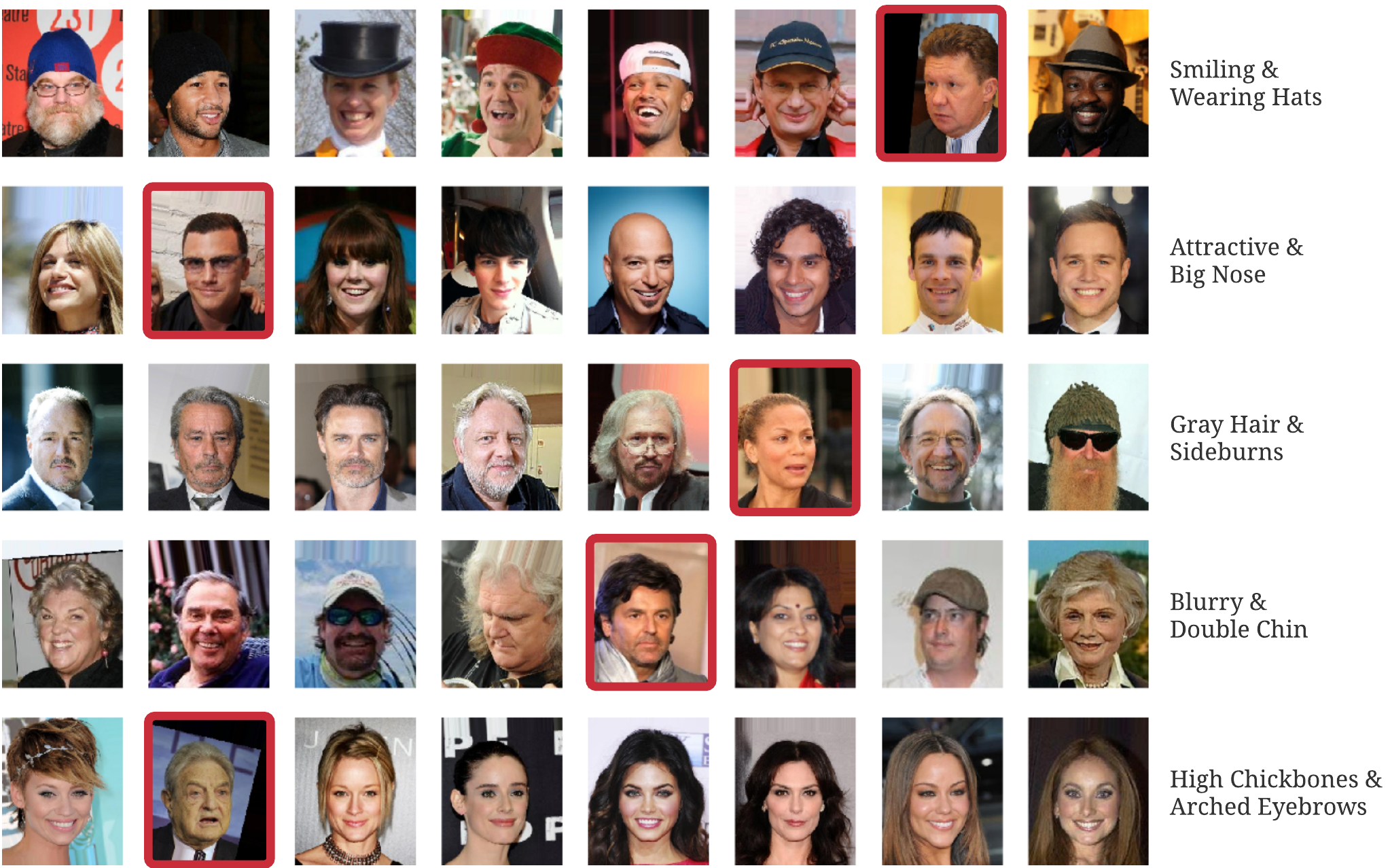}
    \caption{Example input sets for anomaly detection. Each row is a set containing 7 images with two attributes present and one image with both absent. The anomalous image is shown in a red frame.} \vspace{-4mm}
    \label{fig:my_label}
\end{figure}
\subsection{Set Anomaly Detection}
Here, we build 1000 sets of 8 images, as described in section \ref{exp} and use 800 sets for training. First, we cropped 160$\times$160 patches from images, and then downsample to 64$\times$64 pixels. Images are passed through 9 2D convolutions with and 3 max-pooling layers (one after each set of three convolution layers), that produce an output with the dimension of batch-size.$n\times$256, where $n=8$ is the number of elements in each set. In other words, the input of PICASO will be a set of 8 256D elements. We found this experiment more challenging compared to the previous ones. Hence, we utilize generalized PICASO (GPB) with three updating steps with weight sharing using Eq. \ref{eqg}. Further, we add another layer after PICASO that computes the difference between the updated set and the aggregated vector. It will bring several benefits to our model. First, the model finds the distance between each set element and computed mean in PICASO, and as a result, it can easily find the anomalous image in each set. Second, in this case, the model output will be permutation equivariant since the output should change if the position of the anomalous element in the set is changed. Another important thing is that our model can handle sets with arbitrary sizes and find the probability of each image being anomalous. In this experiment, models are trained with cross-entropy loss function, minimized by the Adam optimizer with a learning rate of $1 \times 10^{-5}$ and a batch size of 8. We use the same training setting across all models.
\begin{table}
\begin{center}
\begin{tabular}{|l|cc|}
\hline
  & \bf AUROC & \bf AUPR \\
\hline\hline
random guess & 0.5 &0.125\\
Set Transformer  &0.5941  &0.4386\\
rFF + PMA &0.5564 &0.1583\\
rFF + SA + PMA &0.6456 &0.2657\\
SA + PMA &0.6872&0.2987\\
\hline
rFF + PB &0.5783&0.1691\\
rFF + SA + PB        &0.7568  &0.4159\\
SA + PB       &0.7928 &0.472\\
SA + GPB   &\bf 0.8100  &\bf 0.5317\\
\hline
\end{tabular}
\end{center}
\caption{Set anomaly detection
results} \vspace{-5mm}
\label{table4}
\end{table}
Here, we report the area under precision-recall curve (AUPR) and area under receiver operating characteristic curve (AUROC). For the Set Transformer baseline, we are reporting their published results \cite{lee2019set} since we use the same settings in this experiment. The results obtained from the variations of our model are listed in Table \ref{table4}.

\begin{table}
\begin{center}
\begin{tabular}{|l|ccc|}
\hline
  & \bf AP$_\infty$ & \bf AP$_1$ & \bf AP$_{0.5}$\\
\hline\hline
DSPN$_{10}$ &72.8 &59.2 & 39.0 \\
DSPN$_{30}$ &85.2 &81.1 & 47.4\\
Slot Attention &94.3 &86.7 & 56.0\\
Set Transformer  &50.0  &16.2 & 2.0\\
\hline
PB$_{3}$ &89.6 &74.3 & 33.0\\
\hline
\end{tabular}
\end{center}
\caption{State prediction results} \vspace{-4mm}
\label{table5}
\end{table}

\subsection{State Prediction}
In this experiment, targets are a set of predictions that describe each object. For instance, an object can be a “large green metal cylinder” at position (-2.93, -1.75, 0.70). Each image can contain three to ten objects. Therefore, the output is a set, and it should be permutation invariant. Further, bounding box information is available, and the model should learn which object corresponds to which set element with the associated properties for each image. This setting makes it different from usual object detection tasks. Here, PICASO is used on top of a CNN encoder with 4 convolution layers and positional embeddings to create a set representation for the objects. The outputs of our module are then passed into an MLP. We use 10 template vectors to describe one or more objects in the image. Note that repetitive templates should be matched and removed in cases where we have less than 10 objects in the image. To find matching between PICASO predictions, the Hungarian algorithm \cite{kuhn1955hungarian} is used. We report the results using the standard average precision (AP) metric. The predicted position coordinates are scaled to [-3; 3], and a prediction is correct if there is a matching object with the same properties within a certain Euclidean distance threshold in the 3d coordinate ({$\infty$} means no threshold is applied and only needs all properties to be correct). Here, we use the same settings as the baselines, see \cite{zhang2019deep, locatello2020object}. Results in Table \ref{table5} shows that PICASO performs better than the Set Transformer and competitively with DSPN baselines on this task even though our method is not specifically tailored to this dataset. 

\section{Conclusions}
In this paper, we propose PICASO, a permutation-invariant attention-based set operator which can learn templates through cascaded blocks. This method is entirely based on attention mechanism and can model high-level dependencies and interactions between set elements. Theoretically, we show that PICASO can implement soft K-means algorithm in the special case, which allows the model to handle common variations in the input. We demonstrate the qualitative results from this operator by visualizing the element importance. We also examine the ability of our model in different tasks and it outperforms other set-input neural networks. Our future work will scale up PICASO to handle extremely large input sets with millions of samples.

{\small
\bibliographystyle{ieee_fullname}
\bibliography{egbib}
}

\end{document}